\documentclass[review]{elsarticle}

\usepackage{lineno,hyperref}

\usepackage[english]{babel}
\usepackage[utf8x]{inputenc}
\usepackage[T1]{fontenc}
\usepackage{graphicx}
\usepackage{comment}
\usepackage{color}
\usepackage{xspace}
\usepackage{algorithm}
\usepackage{algpseudocode}
\usepackage{subcaption}

\usepackage[a4paper,top=3cm,bottom=2cm,left=3cm,right=3cm,marginparwidth=1.75cm]{geometry}

\usepackage{amsmath}
\usepackage[colorinlistoftodos]{todonotes}

\journal{Elesevier Expert Systems with Applications Journal}

\newcommand{\MethodName}{Sommelier\xspace}

\begin{document}

\begin{frontmatter}

\title{Automatic Machine Learning  Derived from Scholarly Big Data}

\author[mymainaddress,mysecondaryaddress]{Asnat Greenstein-Messica\corref{mycorrespondingauthor}}
\ead{asnatm@post.bgu.ac.ilAcknowledgments}
\cortext[mycorrespondingauthor]{Corresponding author}
\author[mymainaddress,mysecondaryaddress]{Roman Vainshtein}
\author[mymainaddress,mysecondaryaddress]{Gilad Katz}
\author[mymainaddress,mysecondaryaddress]{Bracha Shapira}
\author[mymainaddress,mysecondaryaddress]{Lior Rokach}



\address[mymainaddress]{Software and Information Engineering Department, Ben Gurion University of the Negev}
\address[mysecondaryaddress]{Be'er Sheva, Israel}

\begin{abstract}
	One of the challenging aspects of applying machine learning is the need to identify the algorithms
that will perform best for a given dataset. This process can be difficult, time consuming and often requires a great deal of domain knowledge. We present \MethodName, an expert system for recommending the machine learning algorithms that should be applied on a previously unseen dataset. \MethodName is based on word embedding representations of  the domain knowledge extracted from a large corpus of academic publications. When presented with a new dataset and its problem description, \MethodName leverages a recommendation model trained on the word embedding representation to provide a ranked list of the most relevant algorithms to be used on the dataset. 
	We demonstrate \MethodName's effectiveness  by conducting an extensive evaluation on 121 publicly available datasets  and 53 classification algorithms. The top algorithms recommended for each dataset by \MethodName were able to achieve on average 97.7\% of the optimal accuracy of all surveyed algorithms. 
\end{abstract}

\begin{keyword}
\texttt{scholarly big data, algorithm recommendation, word embedding, expert system}
\end{keyword}

\end{frontmatter}

\section{Introduction}

The enormous growth in the creation of digital data has created numerous opportunities. Companies and organizations are now capable of mining data regarding almost every aspect of their activity. However, the ability to collect data far outstrips the ability of organizations to apply analytics or derive meaningful insights. One of the main reasons for this gap is the need for human involvement at multiple points of the data-analysis process and the relative scarcity of individuals with relevant skills (i.e., data scientists).

To address the shortage of skilled individuals, researchers have attempted to automate multiple aspects of the data analysis pipeline. Recent studies in this domain have focused on hyperparameter optimization \cite{thornton2013auto}, feature engineering \cite{katz2016explorekit}, data cleaning \cite{chu2015katara}, and automatic generation of deep neural network architectures \cite{zoph2016neural}. Thornton et al. \cite{autoweka2017} suggest an iterative process for the simultaneous selection of the machine learning algorithm and the optimization of its hyperparameters.


We present \MethodName, a framework for leveraging publicly accessible academic publications and open repositories such as Wikipedia to recommend the most suitable algorithms for previously unseen datasets. Based on the intuition that similarly described problems can be solved using similar algorithms, we designed a framework that extracts terms related to machine learning problems and algorithms from Wikipedia. The extracted terms are used to train a recommender system using word embedding techniques applied on a large amount of publicly accessible academic publications.



We conduct our evaluation on the extensive dataset created by Fernandez et al. \cite{fer14}, which contains an exhaustive evaluation of well over a hundred public datasets and algorithms. Our experiments show that \MethodName is highly effective in recommending top performing algorithms for a given dataset. Moreover, the top algorithm recommended by our approach significantly outperforms the results obtained by applying the Random Forest algorithm, a popular ensemble algorithm which was the best-performing algorithm (on average) in the above mentioned study.

\medskip

\noindent Our contributions are as follows:
\begin{itemize}
	\item We present an expert system for recommending top-performing algorithms to previously unseen datasets. The recommendation is based on a word embedding representation of the domain knowledge automatically extracted from a large corpus of relevant academic publications. Moreover, \MethodName does not require extensive analysis of the data itself. We emulate the way a human would approach the problem by relying on relevant previously published work. \MethodName can also be used as a preliminary step for other iterative algorithm recommendation solutions such as Auto-Weka \cite{autoweka2017}. The effectiveness of the proposed approach is demonstrated empirically on a large corpus of publicly available datasets.
    
    
    \item We propose a framework for the automated construction of a structured knowledge-base on machine learning. This goal is achieved by combining unsupervised keyword extraction from Wikipedia with the vast body of work available in public academic repositories. We demonstrate how this knowledge-base can be used to effectively derive actionable insights for machine learning applications.
\end{itemize} 
\section{Related Work}\label{sec:Related_Work}

\subsection{Knowledge base construction from large scale corpora}

The growth in the amount of data available online -- scholarly and otherwise -- provided a significant boost to various attempts to map this data into structured and semi-structured formats and ontologies. The main drive for this was the challenges faced by practitioners in multiple fields to obtain a sufficient amount of labeled data in their respective fields \cite{gil16}.

The best known publicly available large scale 
corpus is no arguably Wikipedia \cite{denoyer2006wikipedia}, and many projects such as DBpedia \cite{bizer2009dbpedia} 
and Wikidata \cite{vrandevcic2014wikidata} use it as a foundation.
Wikipedia has been used successfully in a large variety of tasks, including entity extraction \cite{gattani2013entity}, query expansion \cite{li2007improving}, query performance prediction \cite{Katz:2014:WQP:2600428.2609553} and ranking of real-world objects \cite{katz2017wikiometrics}. Other examples of a large online dataset are YAGO \cite{suchanek2007yago}, which maps entities and their relations, and Wordnet \cite{miller1995wordnet}.

Another group of algorithms for building knowledge graphs from large scale corpora utilizes an iterative approach. Algorithms from this group rely on the knowledge gathered in previous runs to expand and refine their knowledge base. This group of algorithms includes NELL \cite{mohamed2011discovering} -- which explores relations among noun categories -- and Probase \cite{wu2012probase} -- a taxonomy for automatic understating of text. An additional member of this group was recently proposed by Al-Zaidy and Giles \cite{gil17}, and includes an unsupervised bootstrapping approach for knowledge extraction from academic publications.

\subsection{Information extraction in scholarly documents}
Scholarly publication are an important source of information to researchers and practitioners alike. For this reason, a significant amount of work has been dedicated to the extraction of structured data and entities (tables, figures and algorithms) from academic papers \cite{so5,so6,so7,so8}. For example, \cite{pc12} presents a method for identifying and extracting pseudo-code segments from academic papers in the field of computer science. Given that pseudo code segments are generally accompanied by a caption, the purpose of the code can often be inferred using regular expressions.

More recently, additional approached for algorithm extraction and analysis have been proposed. Seer \cite{algoseer16} proposed an algorithms search engine that leverages both machine learning and a rule-based system for the detection and indexing of code. in \cite{algoeff17}, the authors present an algorithm for extracting both the algorithm discussed in a research paper and its performance. Tuarob \cite{tuarob2016improving} proposes the use of ensemble algorithms for the same task.

In addition to algorithms focused on extracting code, some recent work has focused on a much broader extraction of data. In \cite{wu2014towards}, the authors propose a big data platform for the extraction of a wide array of meta-features including ISBNs, authorships and co-authorships, citation graphs etc. This work, along with others that stem from it \cite{ororbia2015big,osborne2013exploring} could be used to extend our own framework as it currently focused on text extraction from scholarly data repositories.
\medskip

\subsection{Algorithm selection}
The classical meta-learning approach for algorithm recommendation, uses
a set of measures to characterize datasets and establish their relationship to algorithm performance \cite{bra08}.
These measures typically include a set of statistical measures,
information-theoretic measures and/or the performance of simple algorithms referred to as
landmarkers \cite{bra08}. The aim of these methods is
to obtain a model that characterizes the relationship between the given measures and the
performance of algorithms evaluated on these datasets. This model can then be used to  provide a ranking
of algorithms, ordered by their suitability for the task at hand \cite{sm08,bra08}.

Recent studies \cite{autoweka2017,autosk} suggest an iterative process for the simultaneous selection of the machine learning algorithm. AutoWEKA \cite{autoweka2017}, a tool for automatic algorithm and hyperparameters selection, uses a random forest-based SMAC \cite{hu11} for a given performance measure (e.g. accuracy). Other algorithm selection tools include Auto-sklearn \cite{autosk}. Another study [11] calculates dataset similarity through the generation of metafeatures and the application of automatic ensemble construction.

Unlike the meta-learning approach, which requires a large amount of datasets for each dataset cluster to train a machine learning model for algorithm recommendation, \MethodName relies on the scholarly big data papers and can provide effective recommendations even in cases where the available training set is relatively small. Furthermore, the proposed approach does not require extensive analysis of the data itself. We emulate the way a human would approach the problem by relying on relevant previously published work.  Comparing to the relatively time consuming iterative recommendation solutions \cite{autoweka2017,autosk}, \MethodName provides a fast algorithm recommendation, and can also be used as a preliminary step for more resource-heavy solutions \cite{autoweka2017,autosk}. 
\section{Approach}
\label{sec:Approach}

\subsection{Overview}
Our approach builds on recent advances in the field of natural language processing (NLP), where the technique of word embedding has had success in capturing and quantifying fine-grained semantic relationships among terms. We apply this technique to a large corpus of publicly available academic publications in the field of machine learning and use it to implicitly model the relationships among problems and algorithms. We then expand and refine our model by crawling Wikipedia and leveraging its rich metadata structure (namely links and categories). We use the refined model as a recommendation algorithm whose goal is to pair datasets with algorithms.

\begin{figure*}
	\centering
	\includegraphics[width=0.9\textwidth]{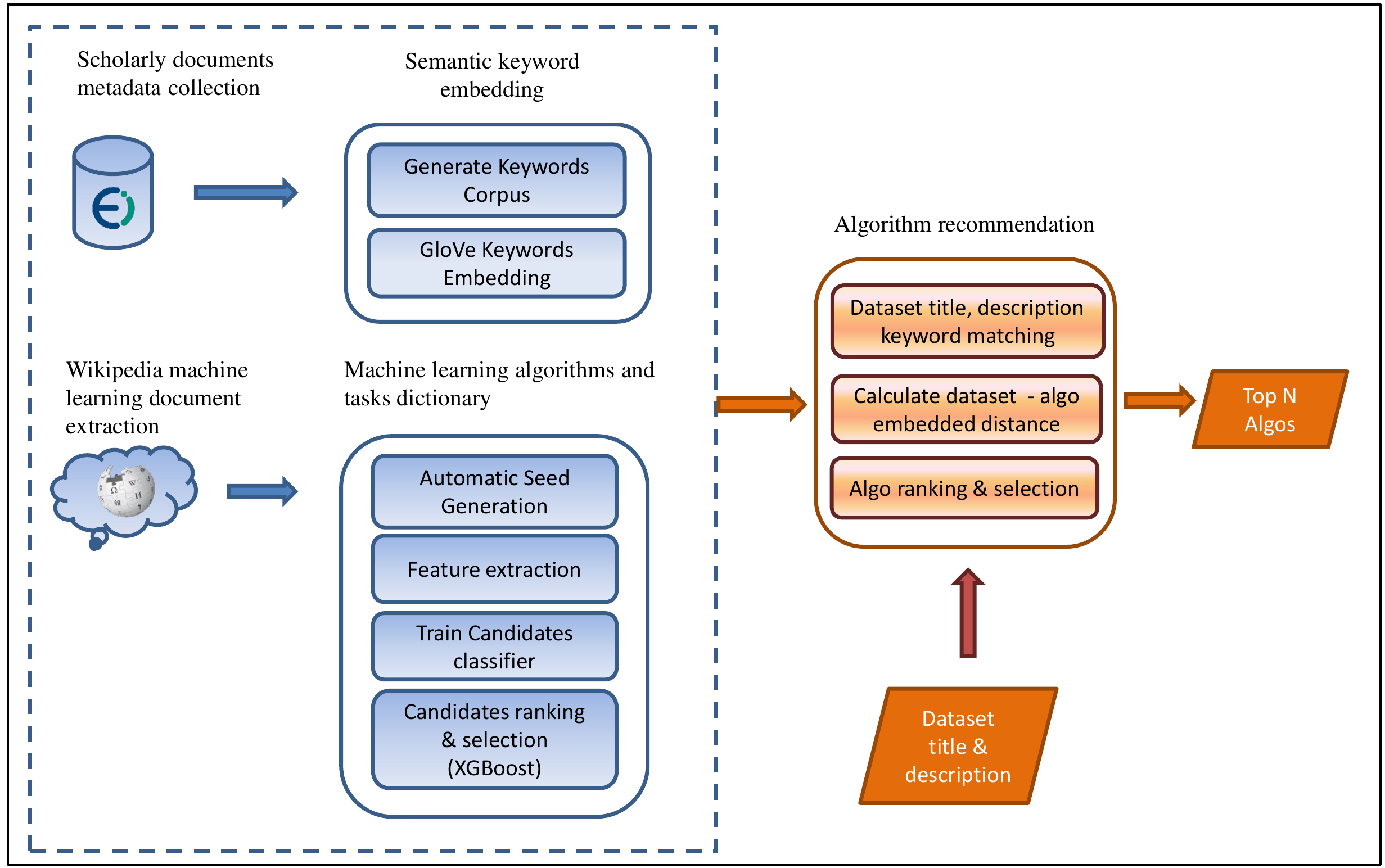}
	\caption{\label{fig:appg}The Sommelier approach pipeline. The blue shapes refer to offline phase, and the orange shapes refer to the online algorithm recommendations for an unseen dataset.}
\end{figure*}

Our approach for recommending algorithms is presented in Figure \ref{fig:appg}. It is comprised of four phases: \textit{corpus extraction}, \textit{semantic embedding generation}, \textit {machine learning-related keyword extraction,} and \textit{recommendation}.

\begin{figure}
	\centering
	\includegraphics[width=0.7\textwidth]{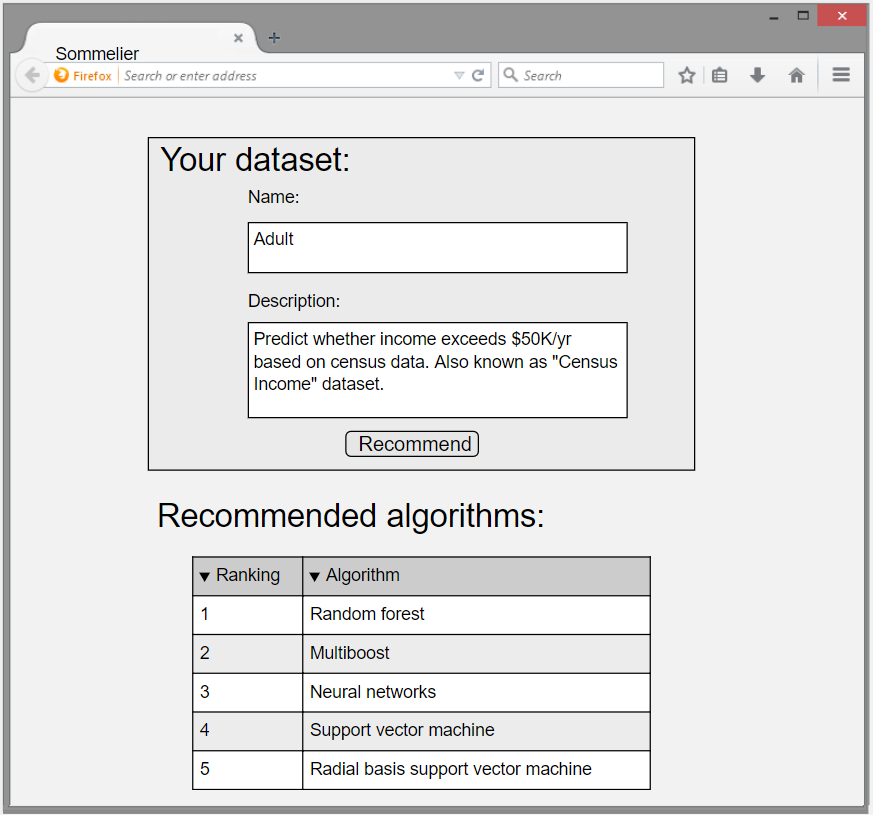}
	\caption{\label{fig:appui}An example of the Sommelier service user interface.}
\end{figure}

During the corpus extraction phase we crawl and retrieve relevant metadata from a large number of machine learning-related papers. The metadata includes features such as the title of the paper, keywords provided by the author and the journal, abstract, publication year, and references to other academic publications. 

In the semantic embedding generation phase we employ GloVe \cite{glove} to create representations of the keywords describing the papers. These representations, after they are refined using data crawled from Wikipedia in the next phase, enable us to identify and recommend algorithms to previously unseen datasets.

In the machine learning-related keyword extraction phase we crawl Wikipedia and use an unsupervised machine learning approach to extract terms related to machine learning algorithms or problems. 
We match these labelled terms to the terms of the embedded representation. By doing so we are able to identify specific implicit connections between ``algorithm'' terms and ``problem'' terms in our embedded representation. 
terms.

During the recommendation phase we receive the title and description of a previously unseen dataset as input. We perform a keyword extraction process, similar to the one used in the machine learning-related keyword extraction phase. This process produces a vector representation of the new dataset, which is then compared with the vectors associated with each of the ``algorithms'' terms in our embedding. Based on the degree of similarity, we produce a ranked list of recommended algorithms. An example of \MethodName's user interface is presented in Figure \ref{fig:appui}. 
\medskip

\noindent The phases of the process are described in detail below.

\subsection{The corpus extraction phase}
\label{subsec:corpusPaperExtraction}
The goal of this phase is to generate a corpus of metadata on machine learning-related papers. To obtain a large number of papers, we crawl the Engineering Village website\footnote{https://www.engineeringvillage.com/home.url} -- a large repository of academic papers which offers access to 13 databases of engineering literature and patents. We applied the following steps to download machine learning-related papers:
\begin{enumerate}
	\item We downloaded all of the papers whose text contained at least one of the following terms: \textit{machine learning}, \textit{data mining}, \textit{regression}, \textit{supervised learning},\textit{ unsupervised learning}, \textit{decision trees}, \textit{boosting}, \textit{random forest}, \textit{neural networks}, \textit{ANN}, \textit{deep learning}, \textit{recurrent neural network}, \textit{RNN}, \textit{convolutional neural network}, \textit{CNN}, \textit{relevance vector machine}, \textit{RVM}, \textit{support vector machine}, \textit{SVM}, \textit{k-means}, \textit{DBSCAN}, \textit{mean-shift}, \textit{bayesian networks,} or \textit{feature engineering}.
	
	\item We downloaded all of the papers that appeared in the following list of top machine learning journals or papers whose citations include papers that appeared in these venues: Data Mining and Analysis\footnote{https://tinyurl.com/y9sgx99x},
	AI\footnote{https://tinyurl.com/yattayqu},
	Computer Vision and Pattern Recognition\footnote{https://tinyurl.com/y8bjlh8j} 
	Database and Information Systems\footnote{https://tinyurl.com/y9l4of7n} and
	Probability and Statistics with Applications\footnote{https://tinyurl.com/yaeuukvx}.
	
\end{enumerate}

Overall, we downloaded the metadata of 461,420 papers, published between 1961 and 2017. For each paper, we stored the paper ID, authors and journal keywords as well as the year of publication. To enable aggregation of similar keywords, we applied standard text normalization on the keywords. The normalization included transforming the text into lower case and replacing space and dash characters with underscores. Following the normalization process we were left with 1,395,788 keywords in our database.

\subsection{The semantic embedding generation phase}
\label{subsec:semanticEmbedding}
The goal of this phase is to generate word embeddings that model the problem--algorithm relationships described in the papers that were extracted in the previous phase. Word embeddings are often used in multiple NLP tasks to discover semantic relatedness among terms \cite{glove,word2vec}. Many studies in this area are based on the distributional hypothesis \cite{word2vec,w2vexp}, which states that words that appear in similar contexts have close meanings. By representing each term as a vector, word embeddings enable us to identify terms with similar meanings even if there are no co-occurrences of the terms in the same document. We hypothesize that this property will enable us to identify effective algorithms for a given problem even if the particular approach has not been previously attempted.

To generate the semantic embedding representations of the papers' keywords, we first needed to create a corpus of candidate keywords. The open--source algorithm, GloVe \cite{glove}, is a highly scalable solution that generates predictive models for unsupervised learning of word embeddings from text. We applied GloVe to all the extracted (normalized) keywords found in the academic papers downloaded from Engineering Village (see Section \ref{subsec:corpusPaperExtraction}). 



GloVe is based on the global log-bilinear regression model and combines the advantages of the global matrix factorization \cite{lee1999learning} and local context window \cite{kawakami2011high}  methods. GloVe explicitly factorizes the word-context co-occurrence matrix on symmetric word windows across the corpus. The embedded word representation is calculated by minimizing the following loss function using gradient descent.

\[J=\sum_{i,j}^{V} f(X_{ij}) (w_i^Tw_j^{~}+b_i+b_j^{~}-log(X_{ij}))^2\]

where $V$ is the number of words in the vocabulary; $X_{ij}$ denotes the number of times word j occurs in the context of word i, while also taking into account the distance between the items within the context window; $w_i$ is the vector representation of word i (i.e., the word embedding), and its size $e$ is the latent embedding size; $w_j^{~}$ is the context item vector, $b_i , b_j^{~}$ are bias terms; and $f$ is a weighting function that cuts off  low co-occurrences, which are usually noisy, as well as prevents overweighting high co-occurrences. The parameters $w_i w_j^{~}, b_i ,$ and $b_j^{~}$ are learned during training. 

To adapt GloVe to our needs we enhanced the co-occurrence factor $X_{ij}$ in the equation with a weight factor to increase the influence of recent papers. The weight factor is equal to 1 for papers published before the year 2000, and increases linearly for later years. The calculation is performed as follows

\[
X_{ij} = \sum_{p \in{P_{ij}}} Y_p
\]

\[
Y_p=
\begin{cases}
1,          \text{if $p$ was published before 2000} &\\
paper\_pub\_year-2000+1, &\text{otherwise}
\end{cases}
\]

\noindent where $P_{ij}$ is the set of papers where keywords $i$ and $j$ co-occur. 

The weighting function $f(X_{ij})$ is designed to reduces the weight of keywords with rare co-occurrence (``noise'' reduction) while also limiting the contribution of common co-occurrences. The weighting function is represented as follows
\[
f(x)=
\begin{cases}
(\frac{x}{x_{max}})^\alpha  &\text{if $ x \leq x_{max}$}\\
1,                         &\text{otherwise}
\end{cases}
\]
where $x$ is the co-occurrence term of two keywords and $X_{max}$ is saturation co-occurrence value.

After filtering out keywords which appeared less than 5 times in the corpus extracted from Engineering Village, we were left with a vocabulary of 120,700 keywords. Our described adaptation of GloVe (including the enhancements to the embedding process) is publicly available\footnote{The link will be added pending acceptance of the paper}. The frequently recommended values of $\alpha=0.75$, and $x_{max}=10$ were used in our experiments. An embedding size $e=200$ with 20 iterations provided good results in our experiments. The process is presented in Algorithm \ref{alg:embedding}. 

\begin{algorithm*}
	\footnotesize
	\caption{Semantic embedding generation}\label{alg:embedding}
	\begin{algorithmic}[1]
		\Procedure{KeywordsEmbedding}{papersMetadataSet, gloveParameters}
		\State $\textit{corpus} \gets \emptyset $
		\For{\textbf{each} (\textit{paper}) in 		\textit{papersMetadataSet}}
		\State $\textit{(paperKeywords,paperPublicationYear)} \gets \textbf{GetPaperMetadata}(paper) $
		\State $\textit{corpus} \gets \textbf{updateCorpus}(paperKeywords,paperPublicationYear) $
		\EndFor
		\State $\textit (autoMLVocab,autoMLVectors) \gets \textbf{GloveML}(corpus, gloveParameters)$       
		\State \textbf{\textit{return}} \textit{(autoMLVocab,autoMLVectors)}
		\EndProcedure
	\end{algorithmic}
\end{algorithm*}
\subsection{The machine learning-related keyword extraction phase}
\label{subseqMachineKeywords}
The goal of this phase is to compile two lists: ``problems'' -- a list of the types of challenges for which machine learning is used and, ``algorithms'' -- a  comprehensive list of machine learning algorithms. Given a new dataset, these two lists will be used to characterize the dataset's traits and recommend relevant algorithms. We hypothesize that our proposed approach can easily take into account all recent and important trends in the field, due to Wikipedia's dynamic and constant update by thousands of contributors.

Our need for generating these two lists -- algorithms and problems
-- stems from the fact that we are unable to know whether a term in the corpus extracted in Section \ref{subsec:corpusPaperExtraction} represents an algorithm, a problem, or neither. By generating these lists from Wikipedia and identifying matching terms, we are able to label relevant terms and filter irrelevant ones. Once the terms are labeled, we can model the algorithm recommendation challenge as a recommendation problem (Section \ref{subsec:RecommendationPhase}). 

The list generation process consists of four phases: \textit{seed generation}, \textit{feature extraction}, \textit{classifier training}, and \textit{candidate ranking and selection}; each phase is described in detail below.

\medskip

\noindent \textbf{Seed generation.} For each of the two types of lists we wish to identify, we first compile the set of page titles that are certain to belong to it:
\begin{itemize}
	\item For the machine learning algorithms, we extracted all of the page titles belonging to the following Wikipedia categories: ``classification algorithms'', ``cluster analysis algorithms'' and ``regression models''. In addition, we extracted all of the algorithms that appeared in the ``machine learning'' bar in the infobox of the machine learning Wikipedia page.
	
	\item For the problems, we extracted the titles of the pages that appeared under ``Applications'' in the infobox of the Machine Learning Wikipedia page.
	
\end{itemize}

\noindent \textbf{Feature extraction.} Next we generate a feature vector to represent every term in the two lists. The vector consists of two types of features:
\begin{itemize}
	\item \textbf{Network-based features.} Since each of our chosen seed terms is represented using a Wikipedia page, we can represent all of the terms in a graph whose vertices are determined by the inter-page links (we construct a single graph containing both lists). For each seed term on either list, we calculated the following values compared to \textit{the seed terms of both lists}: in-degree, out-degree, page rank, betweenness, closeness, hub, authority, and the Dijkstra distance. Each set of values is represented using three statistics: min, max and average.
	
	\item \textbf{Text-based features.} We represent the text of the Wikipedia page corresponding to the seed term using the bag-of-words \cite{joachims1996probabilistic} approach.
\end{itemize}

\noindent \textbf{Classifier training.} After performing the previous steps, we now have two sets of vectors, each representing a single seed term. Next we use these vectors to train a machine learning-based classifier to label previously unseen terms as either as ``algorithm'', ``problem'' or ``other''. To obtain samples for the last label, we randomly sampled Wikipedia pages and labeled them as ``other''. The number of pages belonging to this group was five times the number of pages in the two other groups, combined. \newline

\noindent \textbf{Generating the candidate terms.} In order to expand our lists of algorithms and tasks we first need to identify possible candidates. These candidates will then be classified by the model trained in the previous step. We use three approaches to obtain the candidates:
\begin{enumerate}
	\item We select all Wikipedia articles whose title includes at least one of the following terms: recognition (e.g., speech recognition), analysis (e.g., malware analysis), detection (e.g., plagiarism detection), system (e.g., recommender system, intrusion detection system).
	\item For each seed term on either list, we traverse the Wikipedia graph (constructed based on inter-page links) and retrieve all of the pages that are at most three hops away from a seed concept.
	\item We select all Wikipedia pages whose text contains at least one of the following terms: \textit{machine learning}, \textit{data mining}, \textit{regression}, \textit{supervised learning}, \textit{unsupervised learning}, \textit{decision trees}, \textit{boosting}, \textit{random forest}, \textit{neural networks}, \textit{ANN}, \textit{deep learning}, \textit{recurrent neural network}, \textit{RNN}, \textit{convolutional neural network}, \textit{CNN}, \textit{relevance vector machine}, \textit{RVM}, \textit{support vector machine}, \textit{SVM}, \textit{k-means}, \textit{DBSCAN}, \textit{mean-shift}, \textit{Bayesian networks}, or \textit{feature engineering}. 
\end{enumerate}
Combining these three approaches enabled us to obtain 1.5 million candidate terms.

\medskip

\noindent \textbf{Candidate ranking and selection.}  Next, we apply the trained classifier on the set of candidates. Using the XGBoost algorithm \cite{chen2016xgboost}, we rank all of the candidate terms based on their likelihood of belonging to the ``algorithm'' and ``problem'' labels. For each type, we select the 2,000 top-ranking terms and add them to the relevant set. In order to insure the quality of the newly added terms, we manually review and remove irrelevant terms. The process described above was conducted on an August 2014 version of Wikipedia and resulted in 276 terms describing machine learning algorithms and 380 terms describing relevant challenges.

Finally, following the creation of the two lists we attempt to match the terms on the two list to terms in the embedding. To compensate for small variations in the text (e.g. ``random forest'' and ``random forests''), we use the normalized Levenshtein distance \cite{yujian2007normalized} as the matching criteria. The threshold value for determining a match was set to 0.35. 

\subsection{The recommendation phase}
\label{subsec:RecommendationPhase}
The goal of this phase is to produce a ranked list of  machine learning algorithms with the highest likelihood of being effective for a given problem. We model this challenge as a recommendation problem where our goal is to recommend useful items (algorithms) to users (problems).

The recommendation process begins when we are presented with the title and a short description of the new problem. It is important to note that we do \textit{not} require the actual data to make an effective recommendation (based on \cite{katz2016explorekit}, we do hypothesize  that such information could be useful in future work). We then apply the following steps:

\begin{enumerate}
	\item The dataset title and problem description are normalized and matched with the  vocabulary keywords. The matching of the description is carried out by extracting unigram and bi-gram terms from the text. A list of all of the matched keywords is generated and and each is represented as a vector following the removal of duplicates.
	
	\item Next, we calculate the similarity of each vector generated in the previous section to the algorithms' keyword vocabulary generated in Section \ref{subsec:semanticEmbedding}. For each algorithm vocabulary term, we use cosine similarity \cite{steinbach2000comparison} to calculate its similarity to each dataset title and problem description matched keyword's vector.
	
	\item Each machine learning algorithm in the dictionary is ranked based on the \textit{sum} of its terms' cosine similarity with the terms extracted from the analyzed dataset's title and description. The algorithms are then ranked in  descending order based on their score, using the following equation: 
	\[S_m = \sum_{i \in D} \cos(w_m,w_i)
	\]
	where $w_m$ represents the model keyword embedded vector,  $w_i$ represents the embedded vector for each dataset matched term, and $D$ is the set of all matched dataset keywords.  
\end{enumerate}

\noindent The process is presented in Algorithm \ref{alg:algrec}. The product of this phase is a ranked list of algorithms, sorted by their likelihood of being relevant to the problem at hand. 

\begin{algorithm*}
	\footnotesize
	\caption{Algorithm recommendation}\label{alg:algrec}
	\begin{algorithmic}[1]
		\Procedure{AlgoRecommend}{dataset, autoMLVocab,autoMLVectors,algoKeywordSet,problemKeywordSet}
		\State $\textit{datasetKeywordSet} \gets \textbf{GetDatasetKeywords}( dataset.title,dataset.description,autoMLVocab,problemKeywordSet) $
		\State $\textit{algoDist} \gets 0 $
		\For{\textbf{each} (\textit{datasetKeyword}) in\textit{ datasetKeywordSet}}
		{\For{\textbf{each} (\textit{algoKeyword}) in\textit{ algoKeywordSet}}
			\State $\textit{algoDist.algoKeyword} \gets \textbf{UpdateAlgoDist}(algoKeyword,datasetKeyword,autoMLVectors,algoDist.algoKeyword) $
			\EndFor
		}
		\EndFor
		\State $\textit{recommendedAlgoList} \gets \textbf{RankAlgosByDistance}(algoDist) $ 
		\State \textbf{\textit{return}} \textit{recommendedAlgoList}
		\EndProcedure
		\newline
		\Procedure{UpdateAlgoDist}{algoKeyword, datasetKeyword,autoMLVectors,distance}
		\State $\textit{algoKeywordVec} \gets \textbf{GetVectorForKeyword}(algoKeyword) $
		\State $\textit{datasetKeywordVec} \gets \textbf{GetVectorForKeyword}( datasetKeyword)$
		\State $\textit{keywordsDist} \gets \textbf{CalcCosSimilarity}(algoKeywordVec,datasetKeywordVec) $
		\State \textbf{\textit{return}} \textit{($distance + keywordsDist$)}
		\EndProcedure
		\newline
		
		\Procedure{GetDatasetKeywords}{datasetTitle, datasetDescription,autoMLVocab,problemKeywords}
		\State $\textit{normalTitle} \gets \textbf{NormalizeTitle}(datasetTitle) $
		\State $\textit{normalDescriptionSet} \gets \textbf{SplitAndNormalizeDesc}(datasetDescription) $
		\State $\textit{datasetKeywordSet} \gets \textbf{MatchDatasetToProblems}(normalTitle,normalDescriptionSet,problemKeywords) $
		\If {datasetKeywordSet $\neq \emptyset$ } 
		\State \textbf{\textit{return}} \textit{datasetKeywordSet}
		\Else
		\State $\textit{datasetKeywordSet} \gets \textbf{MatchDatasetToKeywords}(normalTitle,normalDescriptionSet,autoMLVocab) $	
		\EndIf 
		\State \textbf{\textit{return}} \textit{datasetKeywordSet}
		\EndProcedure
	\end{algorithmic}
\end{algorithm*}
\section{Evaluation}\label{sec:evaluation}

We evaluated our approach on the well-known dataset published by \cite{fer14}, which contains the evaluation results of 179 classification algorithms on 121 datasets. The algorithms can be grouped into 17 different ``families'', based on popular criteria. The datasets cover the UCI database in its entirety (as of March 2013, excluding some large-scale problems) in addition to some real-world problems (please see \cite{fer14} for details). For each dataset, all applicable algorithms were applied and evaluated using the accuracy metric. The large scale of the experiments and the diversity of both datasets algorithms ensure that the results were free from collection bias.


The structure of this section is as follows: we first review the models and datasets used in the experiments presented in \cite{fer14} and describe our preprocessing of the data (Section \ref{sec:ExperimentalSetting}). We then present the results of our evaluation (Section \ref{subsec:AccuracyResults}) and analyze the results (Section \ref{subsec:Analysis}). 





\subsection{Experimental setting}
\label{sec:ExperimentalSetting}

In this section we describe the algorithms and datasets included in the evaluation conducted by \cite{fer14}. In addition, we describe the preprocessing steps we applied in order to make sure that the dataset is compatible with the data gathered in Sections \ref{subsec:semanticEmbedding} and \ref{subseqMachineKeywords}. 

\subsubsection{Models}
\label{subsec:Models}
In their evaluation, Fernandez et al. \cite{fer14} used 179 classifiers implemented in C/C++, MATLAB, R, and Weka. The classifiers are highly diverse, originating from 17 ``families.'' A complete list, including the breakdown by family, is presented in Table 1. The main challenge in mapping these algorithms to our embedding was the fact that several algorithms had multiple implementations while our embedding only had a single entry per algorithm (since it is often impossible to infer algorithmic configurations from academic papers). For example, the Random Forest algorithm had eight implementations: \textit{cforest caret, rf\_caret, rforest\_R, parRF\_caret, RRFglobal\_caret, RRF\_caret,} and \textit{RandomForest\_weka}. 

\begin{table*}
	\centering
	\tiny
	\begin{tabular}{|p{4cm}|p{10cm}|}
		\hline
		\textbf{Model} & \textbf{Classifier implementation} \\ \hline
		Adaboost & adaboost\_R, AdaBoostM1\_weka, AdaBoostM1\_J48\_weka, C5.0\_caret\\ \hline
		Adaptive & gcvEarth\_caret, mars\_R\\ \hline
		Bagging & Bagging\_IBk\_weka, Bagging\_RandomForest\_weka, ctreeBag\_R, Bagging\_weka, Bagging\_DecisionTable\_weka, treebag\_caret, Bagging\_PART\_weka,  Bagging\_RandomTree\_weka,  Bagging\_Logistic\_weka, bagging\_R, svmBag\_R, Bagging\_LibSVM\_weka, Bagging\_J48\_weka, ldaBag\_R, plsBag\_R, nbBag\_R, Bagging\_NaiveBayes\_weka,  Bagging\_OneR\_weka, Bagging\_HyperPipes\_weka, nnetBag\_R, Bagging\_DecisionStump\_weka, Bagging\_LWL\_weka, Bagging\_MultilayerPerceptron\_weka\\ \hline
		Bayes net & BayesNet\_weka\\ \hline
		Cascade correlation neural network & cascor\_C\\ \hline
		Decision table & DTNB\_weka, DecisionTable\_weka\\ \hline
		Decision tree & ctree\_caret, RandomSubSpace\_weka, rpart\_caret, REPTree\_weka, rpart\_R, rpart2\_caret, obliqueTree\_R, J48\_weka, J48\_caret, PART\_caret, C5.0Tree\_caret, PART\_weka, NBTree\_weka, ctree2\_caret, RandomTree\_weka, DecisionStump\_weka\\ \hline
		Discriminant analysis & sda\_caret\\ \hline
		Elm neural network & elm\_kernel\_matlab, elm\_matlab\\ \hline
		Ensemble & Decorate\_weka, RandomCommittee\_weka, OrdinalClassClassifier\_weka,     Dagging\_weka, MultiScheme\_weka, Grading\_weka, Vote\_weka\\ \hline
		Flexible discriminant analysis & fda\_caret, fda\_R\\ \hline
		Gaussian kernel & gaussprRadial\_R\\ \hline
		Generalized linear models & glm\_R, mlm\_R, glmStepAIC\_caret, glmnet\_R\\ \hline
		Learning vector quantization & lvq\_caret, lvq\_R\\ \hline
		Linear discriminant analysis & lda\_R, lda2\_caret, PenalizedLDA\_R, slda\_caret, rrlda\_R, stepLDA\_caret, sddaLDA\_R, sparseLDA\_R\\ \hline
		Logistic regression & Logistic\_weka, SimpleLogistic\_weka\\ \hline
		Logitboost & RacedIncrementalLogitBoost\_weka, LogitBoost\_weka, logitboost\_R\\ \hline
		Learning vector quantization neural networks & lvq\_caret, lvq\_R\\ \hline
		Majority voting & Vote\_weka\\ \hline
		Mars & mars\_R\\ \hline
		Mixture discriminant analysis& mda\_R, mda\_caret\\ \hline
		Multiboost & MultiBoostAB\_REPTree\_weka, MultiBoostAB\_DecisionTable\_weka, MultiBoostAB\_MultilayerPerceptron\_weka, MultiBoostAB\_LibSVM\_weka,  MultiBoostAB\_RandomTree\_weka, MultiBoostAB\_Logistic\_weka, MultiBoostAB\_PART\_weka, MultiBoostAB\_RandomForest\_weka, MultiBoostAB\_J48\_weka, MultiBoostAB\_NaiveBayes\_weka, MultiBoostAB\_IBk\_weka, MultiBoostAB\_weka, MultiBoostAB\_OneR\_weka\\
		Multinomial logistic regression & multinom\_caret\\ \hline
		Naive bayes & NaiveBayesSimple\_weka, NaiveBayesUpdateable\_weka, naiveBayes\_R,     NaiveBayes\_weka\\
		Nearest neighbors & knn\_R, knn\_caret, IBk\_weka, IB1\_weka, NNge\_weka\\ \hline
		Neural networks & MultilayerPerceptron\_weka, pcaNNet\_caret, nnet\_caret, avNNet\_caret, mlp\_C,     mlp\_caret, mlp\_matlab, mlpWeightDecay\_caret\\ \hline
		One R & OneR\_weka, OneR\_caret\\ \hline
		partial\_least\_squares\_regression & pls\_caret, gpls\_R, widekernelpls\_R, simpls\_R, kernelpls\_R, spls\_R\\ \hline
		pda & pda\_caret\\ \hline
		pipe & HyperPipes\_weka\\ \hline
		pnn & pnn\_matlab\\ \hline
		quadratic\_discriminant\_analysis & qda\_caret, stepQDA\_caret, sddaQDA\_R, QdaCov\_caret\\ \hline
		random\_forest & cforest\_caret, rf\_caret, rforest\_R, parRF\_caret, RRFglobal\_caret, RRF\_caret,    RandomForest\_weka\\ \hline
		random\_subspace & RandomSubSpace\_weka\\ \hline
		random\_tree & RandomTree\_weka\\ \hline
		rbf\_neural\_network & rbf\_matlab, rbfDDA\_caret, rbf\_caret, RBFNetwork\_weka\\ \hline
		rda & rda\_R\\ \hline
		rep\_tree & REPTree\_weka\\ \hline
		rotation\_forest & RotationForest\_weka\\ \hline
		rule & Ridor\_weka\\ \hline
		rules & C5.0Rules\_caret, OneR\_weka, OneR\_caret\\ \hline
		sda & sda\_caret\\ \hline
		smo & SMO\_weka\\ \hline
		stacking & Stacking\_weka, StackingC\_weka\\ \hline
		support\_vector\_machine & svmBag\_R, svmLinear\_caret, svmlight\_C, svm\_C, LibSVM\_weka, lssvmRadial\_caret,      svmRadial\_caret, svmRadialCost\_caret, svmPoly\_caret, LibLINEAR\_weka\\ \hline
	\end{tabular}
	 
	\caption{Mapping of classifiers described by \cite{fer14} to models vocabulary keywords}
	\label{Tab:models_mapping}
\end{table*}

We addressed this problem by a manually aggregating the different implementations of the same algorithm. After this aggregation was performed, the original 179 machine learning algorithms presented in \cite{fer14} were mapped to 45 entries in the mapping whose creation is described in Section \ref{subseqMachineKeywords}. This information is presented in full in Table~\ref{Tab:models_mapping}.

\subsubsection{Datasets}
\label{subsec:Datasets}
In their evaluation, Fernandez et al. used 121 datasets. These datasets consisted of most of the UCI repository at that time (March 2013) as well as four additional datasets. For a detailed description of these datasets we refer the reader to their publication \cite{fer14}. 

\begin{figure}
	\centering
	\includegraphics[width=0.5\textwidth]{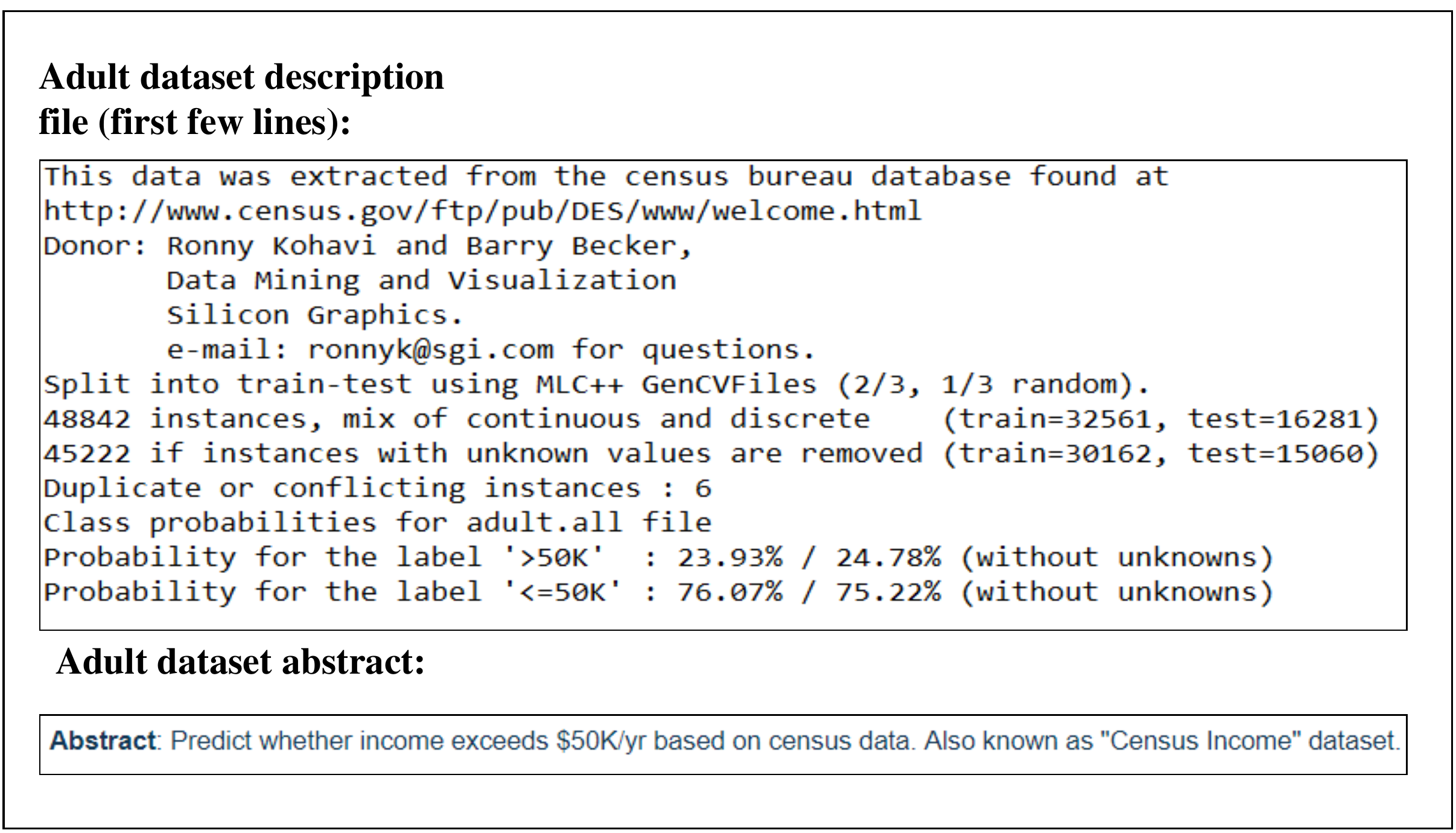}
	\caption{\label{fig:adult}Adult dataset, first few lines of the description file, and the abstract from the UCI website} 
	
\end{figure}

In order to test \MethodName's ability to recommend top-performing algorithms for the datasets described above, we needed the datasets' titles and a short description of their prediction problems. For most of the datasets included in the experiments performed by \cite{fer14}, the authors included an additional file containing a description of the prediction problem as well as a meaningful title. In several cases, though, the problem was not described (please see Figure \ref{fig:adult} which contains the adult dataset description and abstract as an example).

\begin{table*}
	\centering
	\begin{tabular}{c|c|c}
		Measure & RF Relative Accuracy (\%) & Sommelier Relative Accuracy (\%) \\\hline
		Average &  96.4 & 97.7 \\
		Stdev &  7 & 3
	\end{tabular}
	\caption{Relative accuracy of \MethodName vs. Random Forest (RF) across 121 datasets reported by Fernandez et al. \cite{fer14} }
	\label{table:1}
\end{table*}

\begin{table*}
	\centering
	\begin{tabular}{l|c|c}
		Recommendation Type & MRR & Rank Position of Maximum Accuracy Algorithm\\\hline
		Algorithm &  0.28 & 3.5 \\
		Algorithm Family &  0.36 & 2.7
	\end{tabular}
	\caption{\label{tab:MRR}Sommelier recommendation  average MRR, and ranking index of algorithm with maximum accuracy across 121 datasets reported by \cite{fer14} }
	\label{table:2}
\end{table*}

To address this issue, we crawled the UCI website and extracted the abstracts for all of the participating datasets. Then, we combined the UCI dataset's abstract  and description provided by \cite{fer14} when the two were available. We were unable to find a description in the UCI repository for two synthetic datasets (ringnorm and twonorm), and therefore we downloaded this information from the University of Toronto's website\footnote{\href{DELVE repository}{http://www.cs.toronto.edu/delve/data/}}. For the four datasets not included in the UCI repository, we manually extracted the descriptions from the relevant papers. Once the process described above was completed, we were able to assign a title and a description to all of the datasets included in the study. These descriptions were used to rank relevant algorithms, as described in Section \ref{subsec:RecommendationPhase}.

At the end of the process described above, we produce a list of matched keywords for each of the datasets used by Fernandez et al.

\subsubsection{Algorithm performance analysis and comparison}
\label{subsubsec:RankingResults}

One of the conclusions reached by \cite{fer14} was that the Random Forest algorithm, with its different versions, performed best overall. The highest-performing version of the RF algorithm (implemented in R and accessed via caret) achieved an average \textit{relative maximal accuracy} of 94.1\% for all datasets. We define relative maximal accuracy as a percentage of the maximal accuracy obtained by any algorithm for the analyzed dataset. 

To evaluate the effectiveness of our approach, we compared the ranking produced by \MethodName to two baselines. The first is the maximal performance for each dataset, achieved by any algorithm. The second baseline is the performance of the Random Forest algorithm. As it was shown to have the best performance overall (as we explain above), this algorithm is the preferred choice if no information on the analyzed dataset is available.

Because our approach can recommend algorithm types (e.g. Random Forest, logistic regression) but not implementations (e.g. Weka, R) or parameters, for each dataset we \textit{chose the highest performing member} of the relevant algorithm type (please see Table 1 for the complete list).We apply this approach for \MethodName as well as the baselines. For this reason, the average relative maximal accuracy of the Random Forest algorithm is 96.4\% instead of 94.1\%. It is important to note that this setting actually raises the bar for \MethodName compared to the RF baseline.

\begin{figure*}
	\centering
	\begin{subfigure}{0.8\textwidth}
		\includegraphics[width=\linewidth]{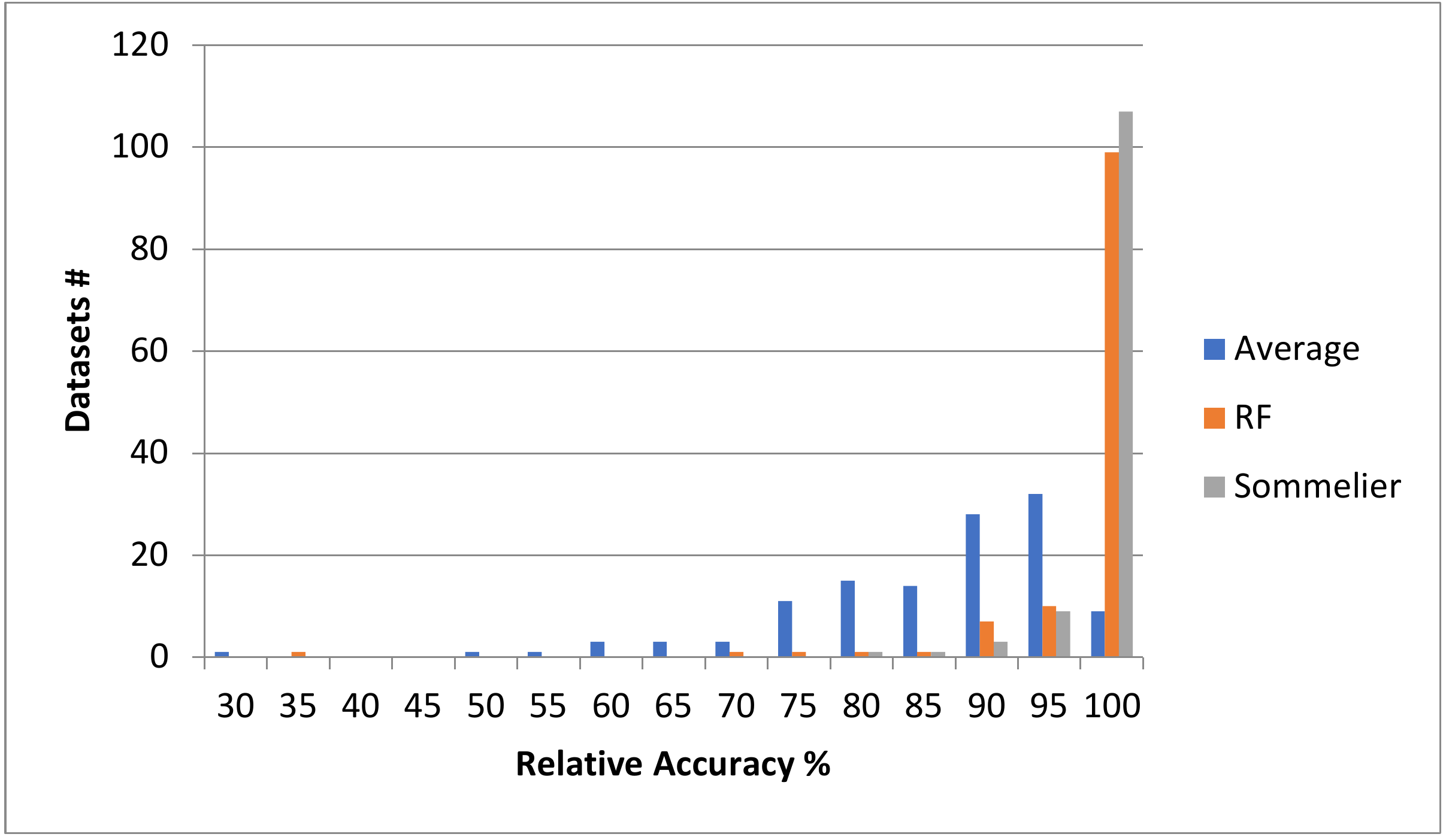}
		\caption{Relative accuracy distribution histogram} \label{fig:1a}
	\end{subfigure}
	\hspace{\fill} 
	\begin{subfigure}{0.8\textwidth}
		\includegraphics[width=\linewidth]{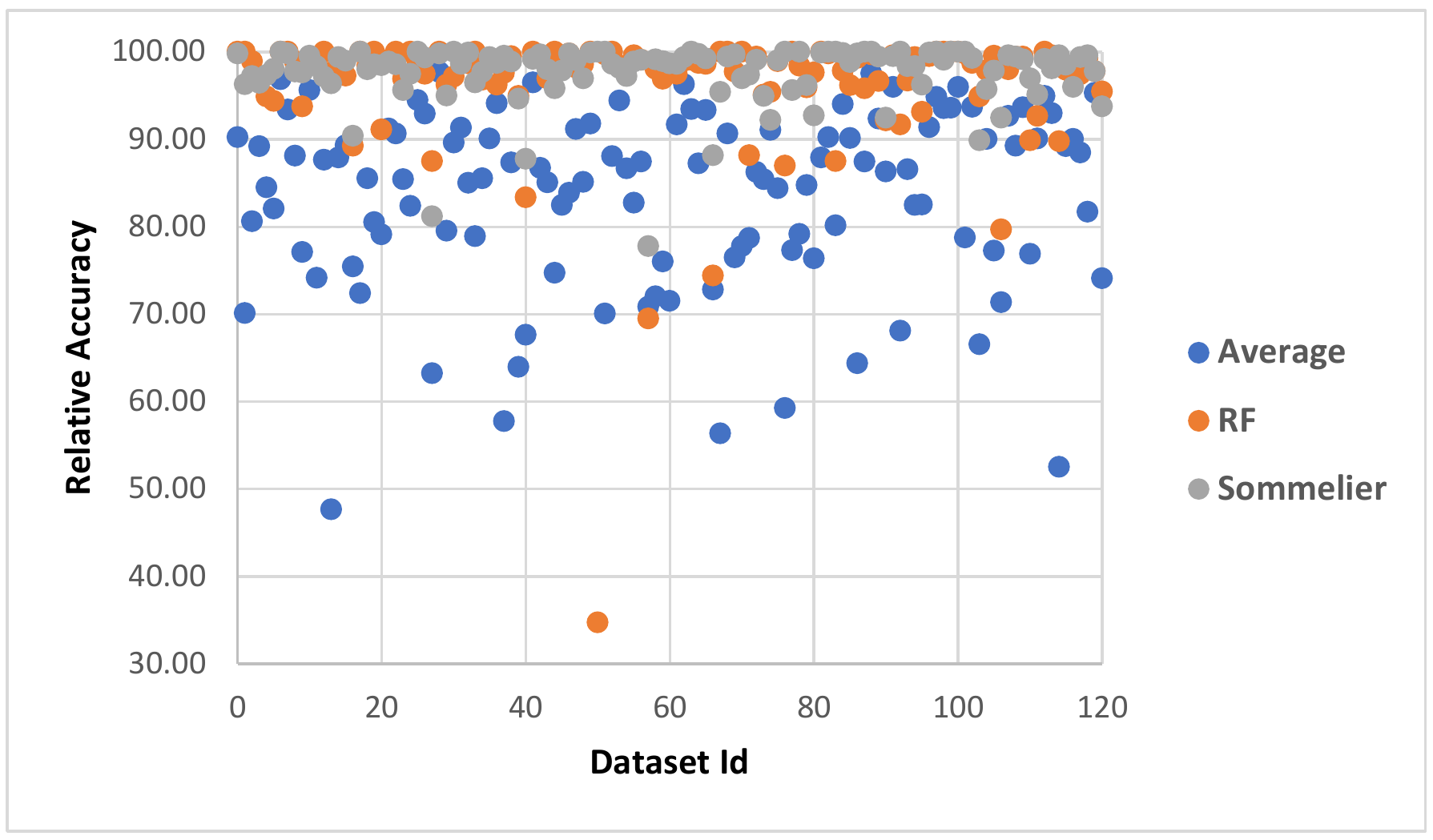}
		\caption{\label{fig:histacc}Relative accuracy distribution} 
	\end{subfigure}
	\caption{\label{fig:datasets}Relative accuracy of 121 datasets for the Sommelier approach, Random Forest algorithm and the average among all algorithms.} \label{fig:1}
\end{figure*}

\begin{figure*}
	\centering
	\includegraphics[width=0.8\textwidth]{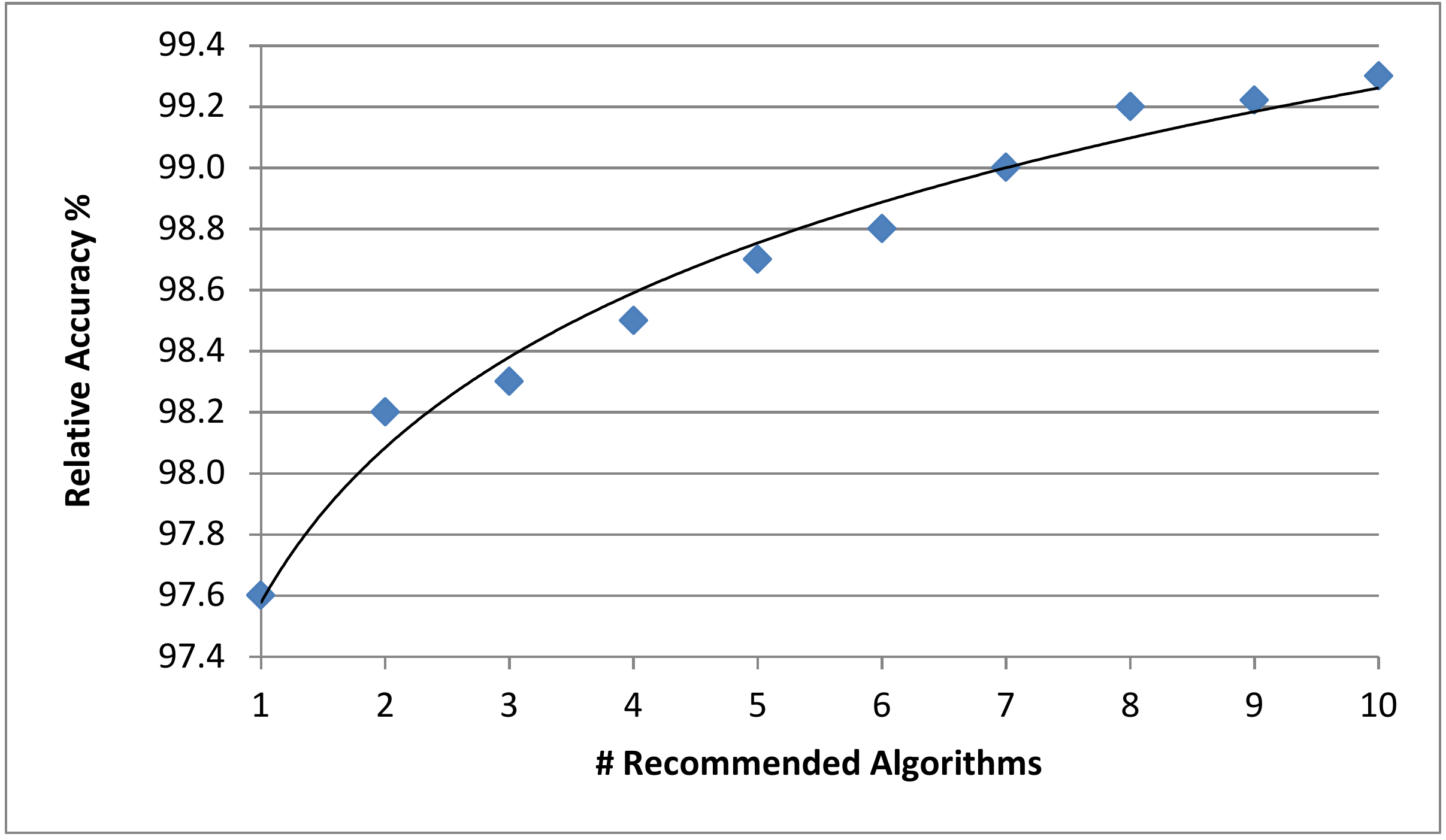}
	\caption{\label{fig:rec_num}Relative accuracy averaged over 121 datasets achieved by Sommelier vs the number of recommended algorithms.}
\end{figure*}

\subsection{Evaluation results}
\label{subsec:AccuracyResults}


The results of our evaluation are presented in Table \ref{table:1}, in which we compare the relative maximal accuracy of both \MethodName and the Random Forest algorithm. The results show that not only does \MethodName outperform the Random Forest algorithm overall but that the performance is consistently closer to the optimal performance (as shown by the lower standard deviation). We evaluated the significance of the results using a paired two-tailed t-test and found the results to be significant with a confidence level of 95\%. 


In Figure \ref{fig:1a} we present a breakdown of the analyzed datasets based on the relative maximal accuracy for \MethodName, the Random Forest algorithms, and an average of all of the algorithms applied to the dataset. The results show that while both \MethodName and the Random Forest algorithm outperform the average overall performance, our approach performs best overall. While both approaches manage to reach a relative maximal accuracy of >95\% in the majority of cases, \MethodName achieved this in 108 of 121 datasets compared to Random Forest's 99. Moreover, the lowest relative maximal accuracy achieved by \MethodName is 77.5\% compared with Random Forest's 35\%. These results indicate that \MethodName is not only more consistent in its performance but that it is mostly able to avoid assigning unsuitable algorithms to a given problem. In Figure \ref{fig:histacc} we compare the performance of \MethodName and Random Forest for each dataset. These results also support our conclusion that while both algorithms fare well overall, the performance by \MethodName is both better and more stable.



\subsection{Analysis}
\label{subsec:Analysis}

\noindent \textbf{Relative maximal performance as a function of the number of evaluated algorithms.} As shown in Table \ref{table:1}, using the top-ranked algorithm by \MethodName would result in an average relative maximal accuracy of 97.7\%. We now explore the effect of evaluating several top-ranked algorithms on this performance measure. The results of this evaluation are presented in Figure \ref{fig:rec_num} and show a consistent increase in average performance. These results lead us to conclude that the ranked lists produced by our approach are effective overall, as they consists of multiple algorithms that achieve high performance for the various datasets.

\noindent \textbf{Number of experiments required to obtain maximal performance.} Next we analyzed the number of algorithms that would have to be evaluated from the ranked list produced by \MethodName in order to obtain the maximal possible performance. To this end we calculated two measures: the relative accuracy as a function of the number of recommended algorithms, and the Mean reciprocal rank (MRR). MMR is a statistic measure used for evaluating the rank of the first correct answer:
\[
MRR = \frac{1}{D}\sum_{i \in{D}} \frac{1}{rank_i}
\]
where $D$ represents the number of datasets evaluated and $rank_i$ is the ranking of the recommended algorithm which match the highest accuracy achieved for the dataset $i$. 

we calculated these two measures for two scenarios: \textit{a)} where each algorithm in the ranked list is evaluated individually (45 possible algorithms) and; \textit{b)} where each algorithm ``family'' puts forward its most effective member (17 possible algorithms). We hypothesize that the latter scenario is of value because some researchers and practitioners may be interested in conducting hyperparameter optimization once an algorithm is selected (using tools such as AutoWEKA \cite{thornton2013auto}). In such cases, the algorithm family is more important than the actual implementation.

The results of our analysis are presented in Table \ref{table:2}. For individual algorithms, one would have to evaluate an average of four algorithms in order to obtain maximal performance. For algorithm families, the number of required evaluations is three. These results also emphasize the advantages of our approach compared with the Random Forest baseline, since the Random Forest algorithm only achieves maximal performance in 18 out of 121 datasets (15\% of cases).

\section{Conclusions and Future Work}
In this study we presented \MethodName, an expert system for recommending which machine learning algorithms should be applied on a previously unseen dataset. When provided with a new dataset, 
our approach analyzes its title and problem and produces a ranked list of algorithms based on their likelihood of performing well on the said dataset. Our approach is based on a word embedding representation of the domain knowledge extracted from a large corpus of academic publications and refined through the use information extracted from Wikipedia. Our evaluation demonstrates that these embeddings can be used to effectively recommend top performing algorithms for diverse datasets with a large variety in size and features composition.

In future work, we plan to incorporate metadata information on the analyzed datasets (when available) into the embedding process. We hypothesize that the metadata can provide additional context and further improve the recommendation accuracy. Examples of such metadata information will include features such as the number of target categories, number of input features and statistical distribution of features.
Furthermore, we plan to extend the process described in this work to include additional types of entities in addition to ``algorithms'' and ``problems''. Such entities may include the performance evaluation metric and type of the used machine learning framework. This expansion of our process can be used to create as automatic machine learning ontology such as \cite{ontology}.

Finally, we plan to explore combining \MethodName with automatic hyperparameter optimization tools such as AutoWEKA \cite{thornton2013auto} or as an initial algorithm recommendation within iterative model selection and hyperparameter optimization tools such as \cite{autoweka2017}.

\section{References}

\bibliography{paperbib}

\end{document}